\documentclass[a4paper]{article}
\usepackage{multirow}
\usepackage{makecell}
\usepackage{INTERSPEECH2022}

\title{NTU Speechlab LLM-Based Multilingual ASR System \\ 
for Interspeech MLC-SLM Challenge 2025}

\name{Yizhou Peng$^1$, Bin Wang$^2$, Yi-Wen Chao$^1$, Ziyang Ma$^{1,3}$, Haoyang Zhang$^{1,4}$\\ 
Hexin Liu$^1$, Xie Chen$^3$, Eng Siong Chng$^1$}
\address{
  $^1$College of Computing and Data Science, Nanyang Technological University, Singapore \\
  $^2$MiroMind, Singapore \quad
  $^3$Shanghai Jiao Tong University, China \quad
  $^4$Peking University, China
}
\email{peng.yizhou@ntu.edu.sg}

\begin{document}

\maketitle
\begin{abstract}
This report details the NTU Speechlab system developed for the Interspeech 2025 Multilingual Conversational Speech and Language Model (MLC-SLM) Challenge (Task I), where we achieved 5th place. We present comprehensive analyses of our multilingual automatic speech recognition system, highlighting key advancements in model architecture, data selection, and training strategies. In particular, language-specific prompts and model averaging techniques were instrumental in boosting system performance across diverse languages. Compared to the initial baseline system, our final model reduced the average Mix Error Rate from 20.2\% to 10.6\%, representing an absolute improvement of 9.6\% (a relative improvement of 48\%) on the evaluation set. Our results demonstrate the effectiveness of our approach and offer practical insights for future Speech Large Language Models.

\end{abstract}

\section{Introduction}

Large Language Models (LLMs) have demonstrated remarkable capabilities in natural language processing, fundamentally reshaping the landscape of artificial intelligence. A burgeoning area of research now focuses on extending these powerful models to the speech modality, leveraging their powerful semantic understanding capabilities to achieve a wider range of functions. This has led to the development of Speech Large Language Models (S-LLMs), which integrate speech encoders, such as Self-Supervised Learning (SSL) models~\cite{hsu2021hubert} and Whisper encoders~\cite{Whisper}, with LLM backbones like Llama~\cite{Llama} and Qwen~\cite{Qwen} to create end-to-end systems for speech processing tasks~\cite{speechgpt, chu2023qwen, xin2024speechtokenizer}, such as Automatic Speech Recognition (ASR). 
    
Recent studies, such as Qwen-Audio~\cite{chu2023qwen}, SpeechGPT~\cite{speechgpt}, and Step-Audio~\cite{step-audio}, have demonstrated the effectiveness of connecting pre-trained audio encoders to large language models (LLMs) through alignment modules. These modules include Q-Formers~\cite{InstructBLIP, BLIVA} and simple linear projectors~\cite{chu2023qwen}, serving to bridge the modality gap between continuous speech representations and the discrete token space of LLMs. Such integration enables the models to leverage the rich linguistic knowledge embedded in LLMs, facilitating the understanding of both linguistic and paralinguistic information encoded in speech.

    

This paper reports the system submitted by NTU Speechlab for \textbf{Track I} of the Interspeech 2025 Multilingual Conversational Speech and Language Model (MLC-SLM) Challenge.
Our contributions are as follows.
\begin{itemize}
    \item We demonstrate that full-parameter tuning of the LLM, combined with a frozen Whisper encoder, is effective for adapting a speech-language model (S-LLM) to automatic speech recognition (ASR) tasks.
    \item We propose to use the language-specific prompts in S-LLMs for ASR, which significantly helps multilingual ASR.
    \item Model Average strategy demonstrates its superior performance gain for an LLM. 
\end{itemize}
Our submitted model achieved \textbf{fifth} place on Track I, significantly improving the baseline performance on the evaluation set in terms of the averaged Mixed Error Rate (MER), from a baseline of 20.2\% to 10.6\%, marking a 48\% relative improvement. 

This report details our system's architecture, strategy, and the experimental results that validate our approach, offering valuable insights for the continued development of high-performance, multilingual S-LLMs.

\begin{figure*}[thb]
    \centering
    \includegraphics[width=0.8\linewidth]{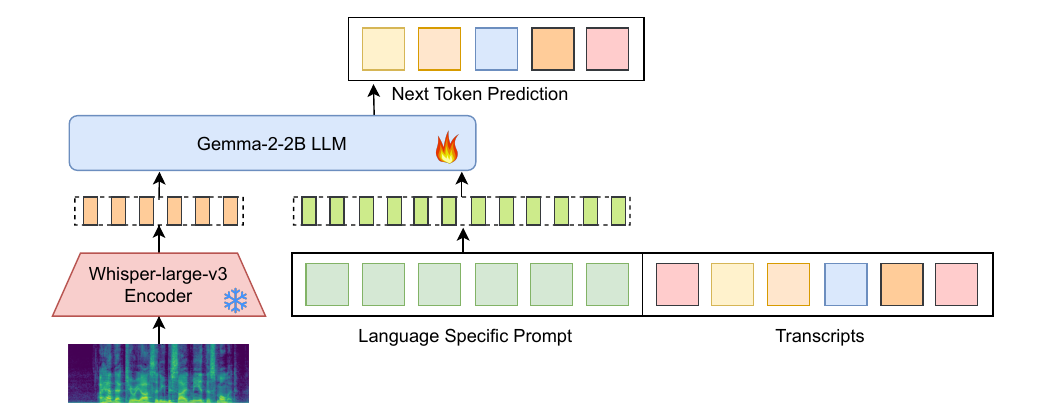}
    \caption{Proposed Model Architecture. In our model, we utilize the Whisper-large-v3 encoder as an audio encoder, and the Gemma-2-2B as the backbone LLM. To understand the audio representation output from the audio encoder, we simply use a linear projector after the encoder. During training, our encoder is frozen, and full-parameter-tuning is applied to both the linear projector and the LLM. The language-specific prompt is specially designed for the MLC-SLM challenge, which uses the prompts shown in Figure~\ref{fig:lid} for each training sample with language label.}
    \vspace{-1.5em}
    \label{fig:model-architecture}
\end{figure*}

\begin{figure}[htb]
    \centering
    \includegraphics[width=0.95\linewidth]{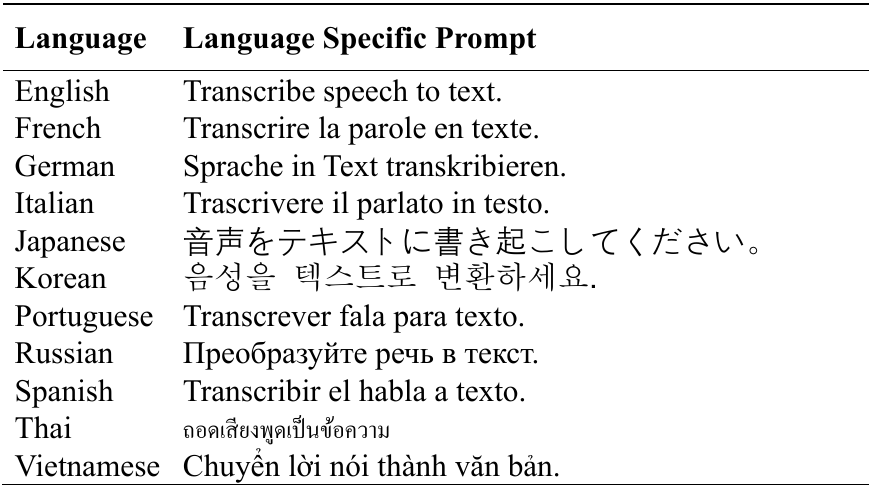}
    \caption{Language Specific Prompt. All these prompts have the same meaning: "Transcribe speech to text," but are written in specific languages based on the language given for a speech.}
    \label{fig:lid}
\end{figure}

\section{Methodology}

In this section, we present the framework of our LLM-based multilingual ASR system. The core design follows a post-alignment structure, where speech features are aligned with language semantic tokens for seamless integration into a pretrained LLM. The model structure is shown in Figure~\ref{fig:model-architecture}. The model transcribes speech input using a language-specific prompt that aligns with the spoken language, ensuring it operates purely as a multilingual ASR system rather than performing translation. The architecture consists of three core components: a speech encoder, a modality adaptor, and an LLM backbone.
    
\textbf{Speech Encoder.} We employ the encoder of Whisper-large-v3 within our SLLM due to its strong multilingual capability, backed by pretraining on diverse linguistic data, to transform speech signals into fixed-length embeddings, covering up to 30 seconds of audio. 

\textbf{Modality Adaptor}. We employ a lightweight two-layer MLP as the adaptor to bridge the speech modality features and the LLM’s embedding space. The adaptor is randomly initialized and trained jointly with the LLM.

\textbf{LLM Backbone}. Unlike monolingual ASR, multilingual ASR demands a more robust language model capable of capturing language-specific patterns essential for accurate transcription. To this end, we employ \texttt{Gemma-2-2B} as the LLM for transcription generation.

We propose to use language-specific prompts during both training and inference, as illustrated in Figure~\ref{fig:lid}. This approach enhances intra-model consistency during autoregressive generation. During supervised finetuning, we freeze the encoder while optimizing the adaptor and LLM via an autoregressive loss. 
Following traditional ASR practices~\cite{espnet, mamba_in_speech}, we further enhance performance through model averaging, applying equal-weighted averaging over the last 15 checkpoints.

\section{Experiments}
In this section, we present the dataset we utilize and the modeling configurations in detail.

\subsection{Dataset}

We train our models on a comprehensive multilingual corpus totaling approximately \textbf{17500 hours} of speech data, including augmentations. The primary resource is the \textbf{MLC-SLM Dataset}, officially provided by the MLC-SLM Challenge. The MLC-SLM dataset contains roughly \textbf{1500 hours} of conversational speech spanning 11 languages: English, French, German, Italian, Portuguese, Spanish, Japanese, Korean, Russian, Thai, and Vietnamese. The English subset includes five accents, including American, British, Filipino, Australian, and Indian, each contributing approximately \textbf{100 hours}. All recordings were collected in quiet indoor environments using consumer-grade devices to ensure high audio quality and realistic conversational conditions.
\begin{table}[htb]
    \centering
    \caption{MLC-SLM dataset statistics. It includes a 1500-hour training set and 32 hours of validation and evaluation sets, covering eleven languages and five different accents in English.}
    \footnotesize
    \begin{tabular}{c|l|c|l}
        \toprule
        Subset & Language & Duration & Notes\\
        \midrule
        \multirow{11}*{Train} & English & 500 & \makecell[l]{\footnotesize 100 hours for each of \\ \footnotesize American, British, \\ \footnotesize Filipino,  Australian, \\ \footnotesize and Indian Accents.} \\
        \cmidrule{2-4}

         & French & 100 & \\
         & German & 100 & \\
         & Italian & 100 & \\
         & Japanese & 100 & \\
         & Korean & 100 & \\
         & Portuguese & 100 & \footnotesize in Europe \\
         & Russian & 100 & \\
         & Spanish & 100 & \footnotesize in Spain \\
         & Thai & 100 & \\
         & Vietnamese & 100 & \\
         \midrule
         Valid & \makecell[l]{\footnotesize All Languages \\ \footnotesize in Train set} & 32 & \makecell[l]{\footnotesize Roughly averaged \\ \footnotesize among languages.}\\
         \midrule
         Eval & \makecell[l]{\footnotesize All Languages \\ \footnotesize in Train set} & 32 & \makecell[l]{\footnotesize Roughly averaged \\ \footnotesize among languages.}\\
         \bottomrule
    \end{tabular}
    \vspace{-1.5em}
    \label{tab:indomain}
\end{table}

To further enhance our multilingual ASR performance, we supplement our training with the following publicly available datasets:

\begin{itemize}
\vspace{-0.2em}
\item \textbf{CommonVoice 21.0}~\cite{commonvoice}: A crowdsourced multilingual corpus of read speech curated by Mozilla. We select the subsets corresponding to the 11 target languages, amounting to approximately \textbf{4467 hours} of audio.
\vspace{-0.2em}
\item \textbf{GigaSpeech}~\cite{gigaspeech} and \textbf{GigaSpeech2}~\cite{gigaspeech2}: GigaSpeech is an English-centric, multi-domain speech corpus collected from podcasts, audiobooks, and YouTube. Its extension, GigaSpeech2, includes additional data for lower-resource languages. We incorporate \textbf{901 hours} of English from GigaSpeech, along with \textbf{2147 hours} of Thai and Vietnamese from GigaSpeech2.
\vspace{-0.2em}
\item \textbf{Multilingual LibriSpeech}~\cite{pratap2020mls}: A large-scale corpus of read speech derived from public domain audiobooks in eight European languages. Our training set includes the French, German, Italian, Portuguese, and Spanish subsets, totaling \textbf{4367 hours} of aligned audio and transcriptions.
\vspace{-0.2em}
\item \textbf{Multilingual TEDx}~\cite{salesky2021mtedx}: A curated speech corpus constructed from TEDx talks on YouTube, featuring time-aligned transcriptions across a wide range of languages. We include French, German, Italian, Portuguese, Russian, and Spanish, contributing approximately \textbf{559 hours} of semi-spontaneous transcribed speech.
\vspace{-0.2em}
\item \textbf{ReazonSpeech}~\cite{fujimoto2016reazonspeech}: A Japanese corpus with around \textbf{486 hours} of transcribed speech from diverse domains such as news, podcasts, and read materials.
\vspace{-0.2em}
\item \textbf{Zeroth-Korean}\footnote{https://github.com/goodatlas/zeroth} and \textbf{Seoul Corpus}~\cite{yun2015korean}: Zeroth-Korean provides \textbf{51.6 hours} of high-quality read speech recorded from 105 native speakers in controlled settings, while the Seoul Corpus offers \textbf{22 hours} of spontaneous Korean speech, including dialogues, monologues, and read passages from speakers of diverse age groups and dialectal backgrounds.
\vspace{-0.2em}
\end{itemize}
The MLC-SLM dataset construction statistics are summarized in Table~\ref{tab:indomain}, and a detailed breakdown of the external dataset composition across languages is provided in Table~\ref{tab:external}.

\begin{table}[ht]
  \centering
  \caption{\textbf{External} open accessible datasets used in model training. The duration is shown in hours. These comprise more than 13k hours in total.}
  \footnotesize
  \begin{tabular}{lll}
    \toprule
    \textbf{Language} & \textbf{Dataset} & \textbf{Duration} \\
    \midrule
    \multirow{2}{*}{English}   
 & Commonvoice 21.0~\cite{commonvoice} & 1786   \\
 & Gigaspeech~\cite{gigaspeech} & 901    \\
 \midrule
    \multirow{3}{*}{French}    
 & Commonvoice 21.0 & 841    \\
 & Multilingual LibriSpeech~\cite{pratap2020mls}   & 1076   \\
 & Multilingual TEDx~\cite{salesky2021mtedx}  & 156    \\
 \midrule
    \multirow{3}{*}{German}    
 & Commonvoice 21.0 & 957    \\
 & Multilingual LibriSpeech & 1966   \\
 & Multilingual TEDx  & 9 \\
 \midrule
    \multirow{3}{*}{Italian}   
 & Commonvoice 21.0 & 254    \\
 & Multilingual LibriSpeech & 247    \\
 & Multilingual TEDx & 79  \\
 \midrule
    \multirow{2}{*}{Japanese}  
 & Commonvoice 21.0 & 20  \\
 & ReazonSpeech~\cite{fujimoto2016reazonspeech}   & 486    \\
 \midrule
    \multirow{3}{*}{Korean}    
 & Commonvoice 21.0 & 1 \\
 & Zeroth-Korean & 51.6   \\
 & Seoul corpus \cite{yun2015korean} & 22  \\
 \midrule
    \multirow{3}{*}{Portuguese}
 & Commonvoice 21.0 & 26  \\
 & Multilingual LibriSpeech & 161    \\
 & Multilingual TEDx & 127    \\
 \midrule
    \multirow{2}{*}{Russian}   
 & Commonvoice 21.0 & 38  \\
 & Multilingual TEDx  & 42.5   \\
 \midrule
    \multirow{3}{*}{Spanish}   
 & Commonvoice 21.0 & 505    \\
 & Multilingual LibriSpeech  & 917    \\
 & Multilingual TEDx  & 146    \\
 \midrule
    \multirow{2}{*}{Thai} 
 & Commonvoice 21.0 & 37  \\
 & Gigaspeech2~\cite{gigaspeech2}  & 1026   \\
 \midrule
    \multirow{2}{*}{Vietnamese}
 & Commonvoice 21.0 & 2 \\
 & Gigaspeech2  & 1148   \\
    \bottomrule
  \end{tabular}
  \vspace{-1.5em}
  \label{tab:external}
\end{table}


\subsection{Experimental setup}

In this challenge, we build four models following the same architecture we propose in Figure~\ref{fig:model-architecture}, as shown in Table~\ref{tab:model_training_details}, apart from the baseline system. 
\begin{figure}
    \centering
    \includegraphics[width=0.90\linewidth]{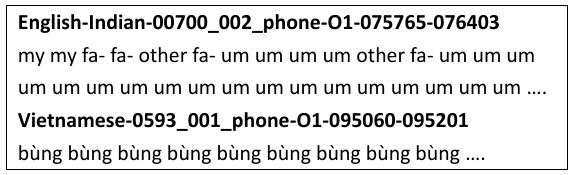}
    \caption{Two examples from the Valid set of hallucinations during the inference period.}
    \label{fig:hallucinations}
\end{figure}
Firstly, the baseline system is provided by the challenge official repo.\footnote{https://github.com/mubingshen/MLC-SLM-Baseline/tree/main/examples/mlcslm/asr} This model follow a two-stage training strategy: first, only the projector is trained, while both the Whisper encoder and the LLM remain fully frozen, to learn a stable mapping from speech embeddings into the LLM’s input space; then, in the second phase, the projector and LoRA adapter of the LLM are jointly fine-tuned, to adapt the language model’s internal representations to speech-derived inputs with minimal tunable parameters.

\begin{table*}[ht]
  \centering
  \caption{Model training configurations. \textbf{Baseline} is released by the challenge official. \textbf{FPT} stands for full parameter tuning. \textbf{CV} is the CommonVoice 21.0 dataset, and \textbf{Ext.} includes all the external datasets listed in Table~\ref{tab:external}. \textbf{LID} means we use a language specific prompt shown in Figure.~\ref{fig:lid}. Duration is shown in hours.}
  \footnotesize
  \begin{tabular}{clrlcclc}
    \toprule
    \textbf{Model ID} & \textbf{Encoder}& \textbf{Decoder}& \textbf{Stratagy}  & \textbf{LR}   & \textbf{Batch Size} & \textbf{Data}   & \textbf{Duration} \\
    \hline
    Baseline  & Whisper & Qwen2.5-7B & LLM-LoRA  & 1e-4  & 208  & MLC-SLM Train  & 1500 \\
    \midrule
    S1  & Whisper & Gemma-2-2B & LLM-FPT & 5e-5  & 96 & MLC-SLM Train   & 1500 \\
    S2  & Whisper & Gemma-2-2B & LLM-FPT (+LID) & 5e-5  & 160  &    MLC-SLM Train   & 1500 \\
    S3  & Whisper & Gemma-2-2B & LLM-FPT (+LID) & 5e-5  & 320  & ~~~~+ CV   & 6000 \\
    S4  & Whisper & Gemma-2-2B & LLM-FPT (+LID)  & 8e-5  & 640 & ~~~~~~~~+ Aug + Ext. & 17500 \\
    \bottomrule
  \end{tabular}

  \label{tab:model_training_details}
\end{table*}
\begin{table*}[ht]
  \centering
  \caption{Word Error Rate (WER$\downarrow$) and Character Error Rate (CER$\downarrow$) results for each of the models. The results for split languages are for the validation set. Mix Error Rate (MER$\downarrow$) is reported for average performance. \textbf{AVG.} uses the latest 15 checkpoints for the equal-weighted model average, where each checkpoint is trained for 400 steps.}
  \footnotesize
  \begin{tabular}{lcccccccc} 
    \toprule
    \textbf{Language} & \textbf{Whisper} & \textbf{Baseline} & \textbf{S1} & \textbf{S2} & \textbf{S3} & \textbf{S3 AVG.} & \textbf{S4 AVG.} & \textbf{Metric} \\
    \midrule
    English-American & 14.14 & 13.83 & 11.89 & 11.52 & 11.07 & 10.39 & 10.58 & WER \\
    English-Australian & 11.72 & 11.19 & 10.25 & 9.34 & 8.38 & 7.63 & 7.48 & WER\\
    English-British & 10.08 & 11.00 & 8.76 & 9.59 & 8.19 & 7.27 & 7.29 & WER \\
    English-Filipino & 9.20  & 8.06 & 9.48 & 9.32   & 7.83 & 6.91 & 7.48 & WER \\
    English-Indian & 13.96 & 16.87 & 14.90 & 15.86  & 14.34 & 12.44 & 12.16 & WER\\
    French & 28.14 & 25.69 & 20.75 & 17.22  & 18.51 & 14.95 & 14.75 & WER\\
    German & 20.72 & 33.95 & 24.53 & 24.35  & 20.75 & 18.29 & 17.80 & WER\\
    Italian & 17.92 & 23.47 & 20.72 & 17.88  & 14.46 & 12.83 & 13.20 & WER\\
    Japanese  & 21.64 & 34.74 & 24.07 & 17.98  & 19.26 & 15.60 & 15.08 & CER \\
    Korean& 13.80 & 20.77 & 13.19 & 12.02  & 11.45 & 9.44 & 9.18 & CER \\
    Portuguese   & 21.23 & 34.02 & 32.97 & 28.66  & 25.29 & 20.72 & 21.65 & WER\\
    Russian & 17.67 & 18.25 & 19.94 & 20.69  & 17.96 & 13.84 & 13.73 & WER \\
    Spanish & 12.27 & 14.31 & 11.43 & 11.39  & 10.00 & 8.85 & 8.81 & WER \\
    Thai  & 14.49 & 21.67 & 13.10 & 10.57  & 9.92 & 9.14 & 8.53 & CER \\
    Vietnamese   & 27.16 & 21.50 & 19.97 & 20.09  & 15.82 & 13.54 & 13.63 & WER\\
    \midrule
    \textbf{Avg. Valid.} & 16.82 & 21.49 & 16.60 & 14.87  & 13.63 & 11.70 & 11.57 & MER \\
    \textbf{Avg. Eval.} & - & 20.17 & - & 14.25 & 13.01 & 10.84 & 10.58 & MER \\
    \bottomrule
  \end{tabular}
      \vspace{-1.5em}
  \label{tab:results}
\end{table*}
Then, we build our models by replacing the backbone LLM from Qwen2.5 to Gemma-2, where the LLM's parameters are \textbf{fully fine-tuned} together with a two-layer linear projector configured with a downsampling rate of 5, to obtain better performance for the ASR task. 
Our S1 model is trained using the MLC-SLM Training dataset only, and the text prompt is fixed to "Transcribe speech to text" regardless of language. 
S2 model uses the same training data as S1 but utilizes language-specific prompts for each language, as shown in Figure~\ref{fig:lid}. 
Extra CommonVoice data is used in S3, incorporating six thousand hours of training data. Finally, S4 model utilizes the largest scale training data among the four models, including not only all the external data demonstrated in Table~\ref{tab:external}, but also simple data augmentation methods are applied to the MLC-SLM training data, including speed (0.9x and 1.1x) and volume (0.15x $-$ 1.15x) perturbations, making up more than 17 thousand hours in total. 

We built our models using the SLAM-LLM~\cite{SLAM} toolkit, running on 8 NVIDIA H20-96GB GPUs. Under our configuration, each GPU can process a batch of 4 samples, and the batch sizes demonstrated in Table~\ref{tab:model_training_details} are calculated by multiplying the number of GPUs, per-GPU batch size, and the steps of gradient accumulation. 
Also, we employ an early-stop strategy during training, with a tolerance of 2000 training steps, based on the validation accuracy. 
Moreover, for the systems S3 and S4, we use the Model Average strategy, equally averaging the last 15 checkpoints, each with 400 update steps, to obtain models with better robustness.


During the inference period, we use beam search with a beam size of 4, and set no repeated ngrams to 5-gram, to prevent hallucinations that were observed in the validation experiments, as examples shown in Figure~\ref{fig:hallucinations}. These samples have similar characteristics that repeat n-gram phrases dozens of times until the end of the sentence. These situations only appeared in less than 0.05\% of the entire valid set, but contribute to more than 0.8\% of WER.

\section{Results}
In this section, we present and discuss the experimental results comparing baseline systems with our proposed models.

Table~\ref{tab:results} summarizes the Word Error Rate (WER) and Character Error Rate (CER) achieved by our models across eleven languages and five accents on the validation set. In detail, we calculate CER for Japanese, Korean, and Thai, while WER is used for the rest of the languages based on the characteristics of each language. For \textbf{Avg.\ Valid.} and \textbf{Avg.\ Eval.}, we report the averaged Mix Error Rate (MER) on both the validation and evaluation sets to show the overall performance of each model. 

The first two columns compare the off-the-shelf Whisper model and the officially released baseline, and the results are from the official repo. Columns S1–S3 show the performance of single best validation accuracy checkpoints, while “S3 AVG.” and “S4 AVG.” report the model average of the latest 15 checkpoints for S3 and S4 models, respectively.

On average, we observe a clear trend of error reduction from \textbf{Baseline} through \textbf{S3}, with the most significant single gain coming from checkpoint averaging. On the validation set, the \textbf{baseline} system achieves an overall MER of 21.49\%, and the MER is reduced to 16.60\% and further to 14.87\% in \textbf{S1} and \textbf{S2}, by fully finetuning the LLM and introducing the language-specific prompts, respectively. 
Then, using external CommonVoice data, \textbf{S3} reduces MER to 13.63\%, and the model average for \textbf{S3} further improves MER performance to 11.70\%. The \textbf{S4} averaged model yields the best validation performance at 11.57\%, and delivers a consistent improvement on evaluation data (10.58\% MER) compared to \textbf{S3 AVG.} (10.84\% MER). These results confirm that model averaging via an equal-weighted strategy substantially mitigates variance between training steps and enhances generalization.

A language-by-language breakdown highlights that high-resource varieties such as English (American, Australian, British accents) and Spanish attain the lowest absolute WERs, which are under 9\% with \textbf{S4 AVG.}, reflecting ample training data. Languages with richer morphology or tonal distinctions (e.g., French, Portuguese, Vietnamese) start with higher WER (20–35\% from Whisper/Baseline) but still achieve relative reductions of up to 35–45\% by \textbf{S4 AVG.}.  For character-based languages like Japanese, Korean, and Thai, CER improvements of 10-20\% in absolute terms demonstrate the robustness of our approach across diverse writing systems.

The consistent gains from S1 through S4 AVG. underscore two key insights for system design. First, model averaging is a simple yet powerful method to enhance performance without requiring architectural changes, additional data, or further training steps. Second, the use of more external data and data augmentation methods can only improve the model's performance by a limited margin, suggesting that utilizing more out-of-domain data is a sub-optimal choice for achieving better ASR results on a specific domain. 
Overall, our results validate the effectiveness of model averaging and full parameter fine-tuning for multilingual speech recognition.

\section{Conclusion}

In this report, we have presented a comprehensive description of our system developed for Task I of the MLC-SLM challenge, achieving the 5th place overall. Our detailed exploration of model architectures, careful utilization of data, and strategic deployment of language-specific prompts and model averaging proved instrumental in enhancing the ASR performance of LLMs. These insights and approaches have demonstrated clear effectiveness and offer valuable directions for future improvements on Speech LLMs.

\bibliographystyle{IEEEtran}
\newpage
\bibliography{mybib}

\end{document}